\documentclass[runningheads]{llncs}

\usepackage{eccv}

\usepackage{eccvabbrv}

\usepackage{graphicx}
\usepackage{booktabs}
\usepackage{multirow}
\usepackage[accsupp]{axessibility}  

\newcommand{\correspondingauthor}{\textsuperscript{*}}

\usepackage[pagebackref,breaklinks,colorlinks,citecolor=eccvblue]{hyperref}

\usepackage{orcidlink}

\begin{document}

\title{SpikeNVS: Enhancing Novel View Synthesis from Blurry Images via Spike Camera} 

\titlerunning{SpikeNVS: Enhancing Novel View Synthesis from Blurry Images via Spike Camera}

\author{
Gaole Dai\and
Zhenyu Wang \and
Qinwen Xu \and
Ming Lu\and
Wen Chen\and
Boxin Shi\and
Shanghang Zhang \correspondingauthor \and
Tiejun Huang
}


\institute{
National Key Laboratory for Multimedia Information Processing,\\ School of Computer Science, Peking University
}
\maketitle
\footnotetext[1]{\correspondingauthor\ Corresponding author.}
\begin{abstract}
One of the most critical factors in achieving sharp Novel View Synthesis (NVS) using neural field methods like Neural Radiance Fields (NeRF) and 3D Gaussian Splatting (3DGS) is the quality of the training images. However, Conventional RGB cameras are susceptible to motion blur. In contrast, neuromorphic cameras like event and spike cameras inherently capture more comprehensive temporal information, which can provide a sharp representation of the scene as additional training data. Recent methods have explored the integration of event cameras to improve the quality of NVS. The event-RGB approaches have some limitations, such as high training costs and the inability to work effectively in the background. Instead, our study introduces a new method that uses the spike camera to overcome these limitations. By considering texture reconstruction from spike streams as ground truth, we design the Texture from Spike (TfS) loss. Since the spike camera relies on temporal integration instead of temporal differentiation used by event cameras, our proposed TfS loss maintains manageable training costs. It handles foreground objects with backgrounds simultaneously. We also provide a real-world dataset captured with our spike-RGB camera system to facilitate future research endeavors. We conduct extensive experiments using synthetic and real-world datasets to demonstrate that our design can enhance novel view synthesis across NeRF and 3DGS. The code and dataset will be made available for public access.
\end{abstract}

\section{Introduction}
\label{Introduction}
Novel view synthesis methods, such as Neural Radiance Fields (NeRF) \cite{Mildenhall2020NeRF} and 3D Gaussian Splatting (3DGS) \cite{kerbl20233d}, have garnered significant attention due to their remarkable capability learned from multi-view images. The NeRF model predicts color and density from 3D scene coordinates and ray information, subsequently volume-rendered into an image. On the other hand, 3DGS uses the learnable 3D Gaussian to represent the scene and employs the splatting method for rendering. In both methods, achieving sharp NVS heavily relies on the quality of training images. However, RGB cameras are prone to motion blur, compromising the captured images and learning process. Previous approaches, such as NeRF-W \cite{martinbrualla2021nerf} and Deblur-NeRF \cite{Ma2021DeblurNeRFNR}, have attempted to model image degradation factors like blur, occlusion, and illumination changes. Nevertheless, these methods fail to address the inherent limitations of RGB cameras. In contrast, neuromorphic cameras like event and spike cameras inherently capture richer temporal information than RGB cameras. Consequently, a few recent studies have proposed enhancing NVS with neuromorphic cameras.

\begin{figure}[htb]
    \centering
    \includegraphics[width=\columnwidth]{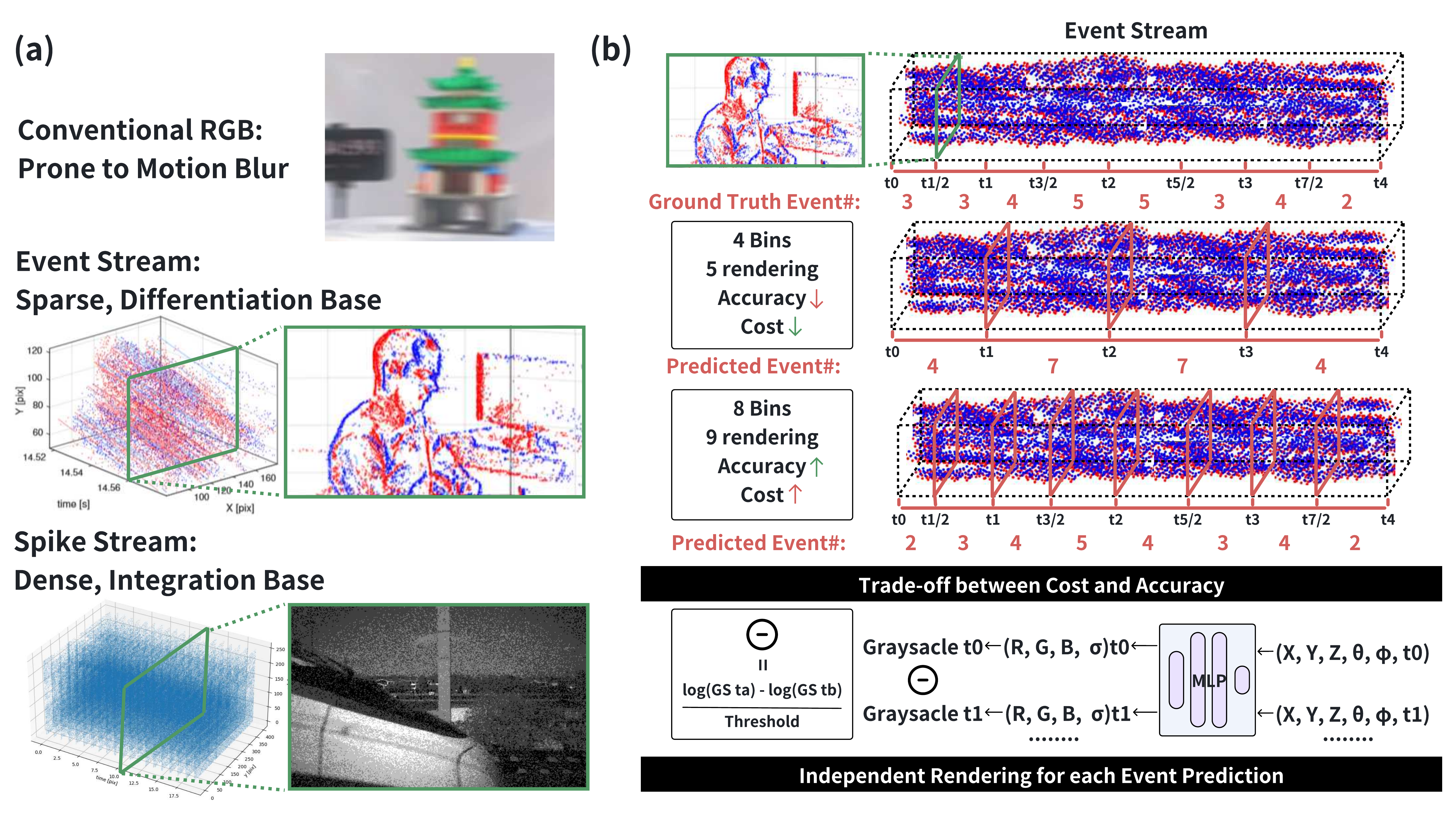}
    \caption{\textbf{Limitations of event-RGB methods.} \textbf{(a)}: Neuromorphic cameras can capture much richer temporal information than conventional RGB cameras. However, event streams are sparser than spike streams and cannot capture static background information.
    \textbf{(b)}: Integrating event streams into NVS methods requires high training costs. Taking E2NeRF as an example, achieving higher accuracy requires increasing the number of independent renderings, resulting in significantly higher costs (top row). The reason for this is that simulating events requires the use of asynchronous differentiation (bottom row).}
    \label{Fig1}
\end{figure}

The primary focus of event-RGB methods is to utilize event cameras for capturing dynamic objects and guiding the learning process \cite{hwang2023evnerf, Qi_2023_ICCV, rudnev2023eventnerf}. However, event cameras rely on temporal differentiation, restricting their ability to capture static scenes like the background. Additionally, these approaches use an event-based loss function that requires multiple independent renderings for differentiation calculation. These methods lead to significantly higher training costs. The limitations shown in Fig.~\ref{Fig1} (b) arise due to the trade-off between precision and computational cost in E2NeRF \cite{Qi_2023_ICCV}, an event-RGB NeRF method.

\vspace{-0.5cm}

\begin{figure}[htb]
    \centering
    \includegraphics[width=1\columnwidth]{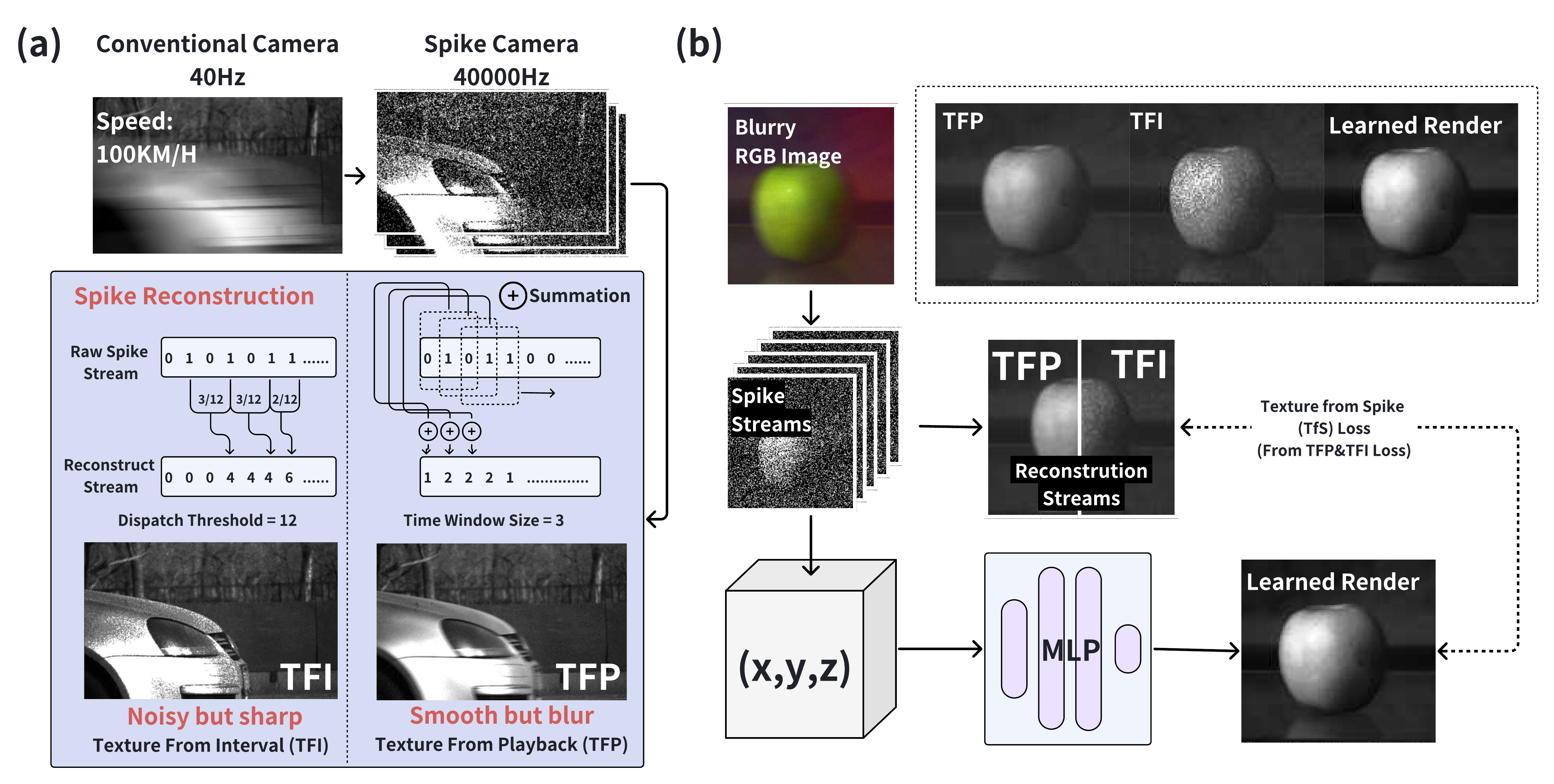}
    \caption{\textbf{Illustration of the proposed TfS loss.} \textbf{(a)}: The Texture from Interval (TFI) and Texture from Playback (TFP) algorithms are commonly employed for spike reconstruction. TFP downsamples along the time axis, while TFI calculates the eligibility trace of each pixel. Both methods encounter a trade-off between noise and sharpness. \textbf{(b)}: The Texture from Spike (TfS) loss combines the losses from TFI and TFP reconstructions, resulting in learned reconstructions that balance noise and sharpness.}
    \label{Fig2}
\end{figure}

In contrast, the spike camera measures accumulated brightness to activate a spike pulse, perfectly avoiding the intrinsic limitations of event-based methods. This temporal integration property allows for the simultaneous capture of static and dynamic scenes, avoiding the need for independent rendering for asynchronous differentiation. With this insight, we propose a novel Texture from Spike (TfS) loss to introduce spike data to the NVS methods. Specifically, Texture from Interval (TFI) and Texture from Playback (TFP) \cite{Dong2017SpikeCA} are two commonly used spike image texture reconstruction techniques. As demonstrated in Fig.~\ref{Fig2} (a), TFI evaluates the relative trigger frequency within an interval and reconstructs the image based on the eligibility traces. Conversely, TFP utilizes a sliding time window (e.g., 32, 64 timestamps) to aggregate the spike pulses within this time window. Our TfS loss leverages the benefits of both TFP and TFI in a learnable manner, as shown in Fig.~\ref{Fig2} (b).

We also build a new dataset with our synchronized spike-RGB camera system to evaluate our method on real-world scenes. In summary, our main contributions can be concluded as follows:
\begin{enumerate}
\item We present a pioneer study that assesses the advantages of spike over event streams regarding both quality and cost for NVS. We have developed the first spike-RGB NVS technique based on this.
\item We propose a novel Texture from Spike (TfS) loss that effectively integrates the advantages of both Texture from Interval (TFI) and Texture from Playback (TFP) results, which are commonly employed for spike reconstruction, into the learning process.
\item We have developed a synchronized spike-RGB camera system with aligned field-of-view and trigger time. We contributed the first real-world dataset for spike-RGB NVS based on it. 
\end{enumerate}

\section{Related Work}
\label{Related Work}
\subsection{Novel View Synthesis}
Neural Radiance Fields (NeRF) have been a popular method for synthesizing novel views since they were first introduced \cite{Mildenhall2020NeRF}. The following advancements in NeRF technology have been directed towards improving its effectiveness and performance. For instance, PlenOctree \cite{yu2021plenoctrees}, FastNeRF \cite{garbin2021fastnerf}, and EfficientNeRF \cite{hu2022efficientnerf} have refined data structures to expedite processing speeds. Other innovations, such as AutoInt \cite{Mildenhall2020NeRF} and Instant-NGP \cite{muller2022instant}, have worked on extracting distinct features to improve scene representation.
Research efforts like pixelNeRF \cite{yu2021pixelnerf}, and MVSNeRF \cite{chen2021mvsnerf} have been made to adapt NeRF to sparse view conditions, leveraging pre-trained networks for feature extraction. Methods incorporating depth or geometry cues \cite{xu2022point, deng2022depth} have been proposed to speed up training. Recently, the emerging technique of 3D Gaussian Splatting (3DGS) \cite{kerbl20233d} employs the learnable 3D Gaussian to represent the scene and utilizes the splatting method for rendering from arbitrary camera views. This approach achieves real-time rendering without using neural networks to learn the implicit function, resulting in accelerated training. However, all these advancements rely on consistent, high-quality multi-view images, which poses a common challenge, especially when using standard RGB cameras.

\subsection{Neuromorphic Cameras}
Spike cameras \cite{dong2021spike} and event cameras \cite{lichtsteiner2006100db, daniel2000mean} are bio-inspired sensors that can overcome the limitations of traditional RGB cameras in challenging scenarios. One of the key advantages of Neuromorphic cameras is their high temporal resolution and pixel bandwidth. Researchers have leveraged these advantages to recover high-quality scene information from low-quality RGB frames. EDI \cite{pan2019bringing} and E-CIR \cite{Song2022E-CIR:} fuse event data and blurry frames to reconstruct a high-quality video. \cite{Zhao2020Motion, Zhao2020High-Speed, Zhao2022Reconstructing} use spike cameras to capture high-speed movements without motion blur. Besides, spike and event cameras have a much higher dynamic range than RGB cameras. This characteristic has been effectively utilized in spike cameras to address challenges such as overexposure \cite{Zhao2021Super}, low illumination \cite{Dong2022High-Speed}, and noise \cite{Zhu2022Ultra-High} associated with conventional cameras. Similarly, event cameras have been used in different ways, such as multi-bracket HDR pipelines \cite{Messikommer2022Multi-Bracket}, E2SRI \cite{Isfahani2022E2SRI:}, and enhancing robustness in dynamic scenes \cite{Isfahani2019Learning}. However, event cameras struggle to capture the texture details of visual scenes due to their recording of only relative changes in light intensity, leading to significantly degraded visibility. In contrast, spike cameras completely record the absolute light intensity at a fairly high frame rate, which provides a more explicit input format for detailed reconstruction.

\section{Method}
\label{Method}
We first commences with a concise introduction to NeRF, 3DGS, and spike reconstruction. Subsequently, we demonstrate of our synchronized spike-RGB camera system. Ultimately, all components are integrated to form our final pipeline.

\subsection{Preliminary}
\label{Preliminary}
NeRF utilizes a Multi-layer Perceptron (MLP) $F_{\theta}$ to model the mapping from 5D input coordinates—combining 3D spatial positions and 2D viewing directions—to the color and density of a scene, as detailed in Eq.~\ref{Eq.1}:
\begin{equation}
    (\textbf{c},\sigma)=F_{\theta} (\gamma(\textbf{o}),\gamma(\textbf{d}))
\label{Eq.1}
\end{equation}
The function $\gamma(\cdot)$ encodes the inputs into a higher dimensional space to facilitate the learning of complex spatial relationships. The rendering process relies on classical volume rendering techniques, which integrate the color and density along camera rays, represented by Eq.~\ref{Eq.3}. Here, $\textbf{r}=\textbf{o}+l\textbf{d}$ defines the ray's path, $\textbf{o}$ is the camera origin, $\textbf{d}$ the viewing direction, and $l$ the distance along the ray.
\begin{equation}
    C(\textbf{r})=\sum_{i=1}^N T_i(1-\exp(-\sigma_i\delta_i))\textbf{c}_i, \quad
    T_i=\exp(-\sum_{j=1}^{i-1}\sigma_j\delta_j) 
\label{Eq.3}
\end{equation}
NeRF leverages hierarchical volume sampling, optimizing both a coarse and a fine network to render accurate images. The loss function, described in Eq. \ref{Eq.4}, minimizes the difference between the rendered and the ground truth colors across a batch of rays \textit{R}, enhancing the fidelity of both the coarse and fine models.
\begin{equation}
    \mathcal{L}=\sum_{\textbf{r}\in\textit{R}}(\| C_c(\textbf{r})-C(\textbf{r})\|_2^2 + \| C_f(\textbf{r})-C(\textbf{r})\|_2^2)
\label{Eq.4}
\end{equation}
3D Gaussian Splatting involves projecting points in 3D space onto the visual plane and smoothly distributing their influence using Gaussian functions. Each point $x$ in the cloud is then represented as a 3D Gaussian. This involves defining parameters such as the position center $\mu$ and covariance $\Sigma$ for each Gaussian. \[G\left ( x \right ) = e^{-\frac{1}{2}\left ( x \right )^{T}\Sigma^{-1}\left ( x \right )}\]
During rendering, given a viewing transformation $W$ the covariance matrix $\Sigma$ in camera coordinates is given as \[\Sigma' = JW\Sigma W^{T}J^{T}\] where $J$ is the Jacobian of the affine approximation of the transformation.

\subsection{Spike Reconstruction}
\label{Spike reconstruction}
The spike camera captures light intensity in a distinct manner, which differs from the exposure method used by RGB cameras and the differential methods employed by event cameras. It integrates light intensity until it reaches a threshold $\Omega$, triggering a spike while keeping the surplus $I$. Given time $t_i$ and accumulator for each pixel as $A$, we have:

\begin{equation}
    A_{t_i}=(A_{t_{i-1}}+I_{t_i})\mod\Omega
\label{Eq.5}
\end{equation}

The pixel's spike value at (x,y) is determined by the accumulator's value and the input, indicating brightness during sampling:

\begin{equation}
    p_{x,y,t_i}=
\begin{cases}
    1, & \text{if } A_{t_{i-1}}+I_{t_i}\ge\Omega\\
    0, & \text{otherwise}
\end{cases}
\label{Eq.6}
\end{equation}

Spike frames are the spikes between interval $t_{i-1}$ to $t_i$, denoted as $F_i$:

\begin{equation}
    F_i=\{\textbf{s}_i(x_i,y_i,p_i)\}_{t_i}
\label{Eq.7}
\end{equation}

The rich data from spike streams are processed to assist NeRF's learning of spatiotemporal textures. Texture reconstruction uses TFI (Texture from Interval) and TFP (Texture from Playback), described by:

\begin{equation}
    \textbf{TFI: }P_{t_i}=\frac{\Omega}{d_{t_i}}, \textbf{TFP: }P_{t_i}=\frac{N_w}{w}*C
\label{Eq.8and9}
\end{equation}

Here, $P_{t_i}$ is the reconstructed texture; for TFI, $d_{t_i}$ represents the temporal interval (latency) between time $t_i$ and the last spike emission. TFI approach is proficient in capturing the contours of textures. For TFP, $w$ denotes the size of the time window and $N_w$ represents the accumulated spike values within this window. By dynamically adjusting the size of the time window according to different contrast levels, the TFP method achieves texture reconstruction with diverse dynamic ranges.

Unlike event data, which only captures motion, TFP and TFI are adept at reconstructing a relatively blur-free texture of both static (e.g. backgrounds) and dynamic (e.g. moving objects) scenes straight from spike streams.

\subsection{Synchronized Spike-RGB Camera System}
\label{Spike-RGB Camera System Setup}
\begin{table}[htb]
\caption{\textbf{The detailed configuration of the spike-RGB camera}. The Vidar spike camera boasts a significantly higher frame rate and dynamic range, enabling it to capture motion and aperture with exceptional clarity and minimal blur. We downsample images from RGB cameras to match the resolution of Vidar.}
\centering
\begin{tabular}{|c|c|c|}
\hline
\multirow{2}{*}{\textbf{Camera Type}} & \multicolumn{2}{c|}{\textbf{Specifications}} \\ \cline{2-3} 
                                       & \textbf{GoPro 9 (RGB)} & \textbf{Vidar (Spike)} \\ \hline
Resolution                             & 1920$\times$1080       & 400$\times$250         \\ \hline
Frame Rate (fps)                       & 120                    & 40000                  \\ \hline
Dynamic Range (dB)                     & 60                     & 100                    \\ \hline
\end{tabular}
\label{Tab1}
\end{table}

\begin{figure}[htb]
    \centering
    \includegraphics[width=\textwidth]{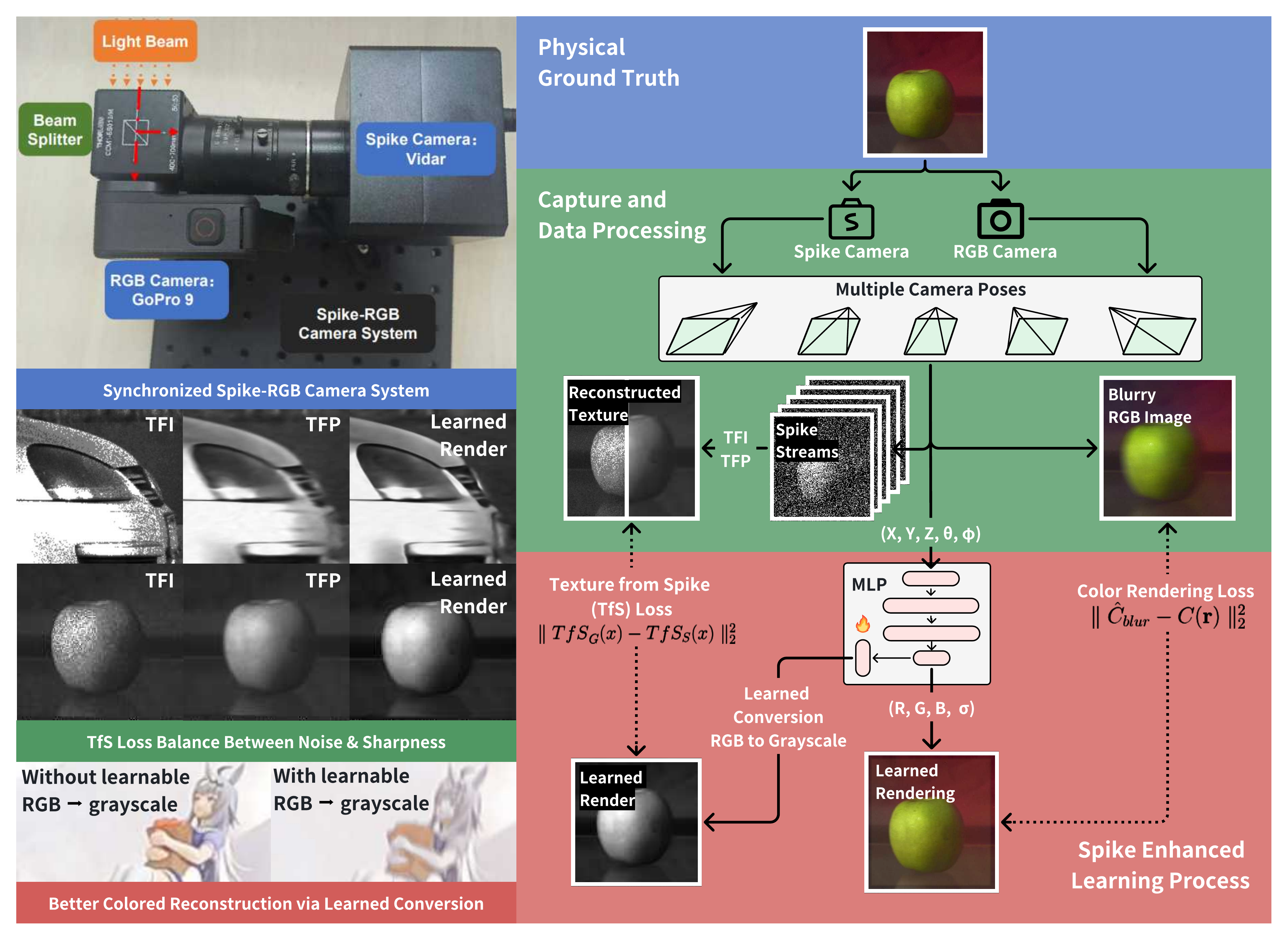}
    \caption{\textbf{SpikeNeRF - Introducing spike streams for NeRF deblur.}  The SpikeNeRF pipeline comprises three stages. To address the issue of blurry RGB images caused by relative motion, we utilize a synchronized spike-RGB camera system to capture spike streams. A detailed configuration of the GoPro9 RGB camera and Vidar spike camera used in this system can be found in Tab.~\ref{Tab1}. The use of a beam splitter ensures a synchronized field of view. Additionally, to mitigate potential misalignment resulting from transmission delays, we incorporate a high-precision clock (0.0001s) at the trigger alignment side as an additional safeguard. On the Texture from Spike (TfS) front, the trade-off between noise and sharpness depicted in Fig~\ref{Fig2} can be effectively balanced during learning by employing both TFP and TFI spike reconstruction techniques. Regarding color rendering, we employ learnable layers to weigh each RGB channel's sum into grayscale, which exhibits greater potential compared to the standard conversion}
    \label{Fig3}
\end{figure}

To introduce spike streams into the learning process of NeRF or 3DGS, the first step is to develop a platform for data collection. In this case, we designed and constructed a synchronized spike-RGB camera system.
This system combines a spike camera (Vidar \cite{dong2021spike}) and a conventional RGB camera (GoPro 9) using a beam splitter (Thorlabs CCM1-BS013). 
The hardware prototype and specifications of our system are depicted in Fig.~\ref{Fig3}.
The purpose of the beam splitter is to achieve the spatial synchronization of the same scene capture.
The beam splitter splits the incoming light and directs it to separate sensors with the same field of view. 
We also ensure time synchronization by employing a clock with an accuracy of 0.0001s to determine the timestamps of the scenes captured by both cameras.
Moreover, the mobility of our Spike-RGB camera system enables us to capture images in both indoor and outdoor environments. This versatility allows us to validate the effectiveness of our proposed method across various scenarios.

\subsection{Integrating Colored Blur-free Representation from Spike Stream}

\paragraph{\textbf{Color Rendering Loss}}
We introduce the concept of Color Rendering Loss as Eq.~\ref{Eq.10}. We adhere to the joint optimization design of NeRF's coarse and fine models, a strategy that remains advantageous in our framework. 
\begin{equation}
    \mathcal{L}_{color}=\sum_{\textbf{r}\in\mathcal{R}}[\parallel\hat{C}_{blur}^c-C(\textbf{r})\parallel_2^2+\parallel\hat{C}_{blur}^f-C(\textbf{r})\parallel_2^2]
\label{Eq.10}
\end{equation}

Here $\hat{C}_{blur}^c$ and $\hat{C}_{blur}^f$ mean the predicted colors of the coarse and fine models, respectively, while $C(\textbf{r})$ denotes the true color values for any sampled ray $\textbf{r}$. 

\paragraph{\textbf{Texture from Spike Loss}}
Our framework enhances deblurring by incorporating spike stream into a learning-based 3D reconstruction process, utilizing our specifically designed Texture from Spike (TfS) Loss. This approach exploits the high temporal resolution of spike cameras, which capture a series of spike streams for each blurry RGB image frame. The TfS Loss aids in reconstructing a clear texture from the spike streams, which corresponds to the non-blurred content within the RGB image's exposure time. We use multiple losses combined to construct the final TfS loss, including the loss from both TFI and TFP. 

The trainable converter layer, which converts the color output into grayscale, is another distinctive design. In contrast to the previous event-based method \cite{Qi_2023_ICCV} that employs standard weighting (R:0.2989, G:0.5870, and B:0.1140) for RGB to grayscale conversion, we question the limited flexibility of such a conversion approach designed for regular images. Instead, we utilize a learnable layer for conversion that aligns the reconstructed grayscale texture with the blur-free texture obtained from spike streams. This design not only demonstrates greater potential but also enables direct loss computation. The formulation of TfS Loss is as follows:
\begin{equation}
    \mathcal{L}_{TfS}=\mathcal{L}_{color} + w\sum_{x\in\mathcal{X}}\|\text{TfS}_{\text{G}}(x) - \text{TfS}_{\text{S}}(x)\|^2_{2}
\label{Eq.TfS}
\end{equation}
In this equation, for any 4D output $x$ from the color rendering head, $\text{TfS}_{\text{G}}(x)$ represents the grayscale texture predicted by the learned conversion, and $\text{TfS}_{\text{S}}(x)$ is the ground truth texture from the spike reconstruction. The overall training loss for SpikeNeRF combines the TfS Loss with a color rendering loss, ensuring that both textural and color are preserved in the deblurred output.

\section{Experiments}
\label{Experiments}
 The First section of the Appendix will provide a more comprehensive account. However, due to spatial constraints, we will present a concise overview on the generation and collection of spike data, experimental setup, and result analysis in this section to showcase our contributions.
\subsection{Experimental Setups}
\label{Datasets}
\begin{figure}[htb]
    \centering
    \includegraphics[width=\textwidth]{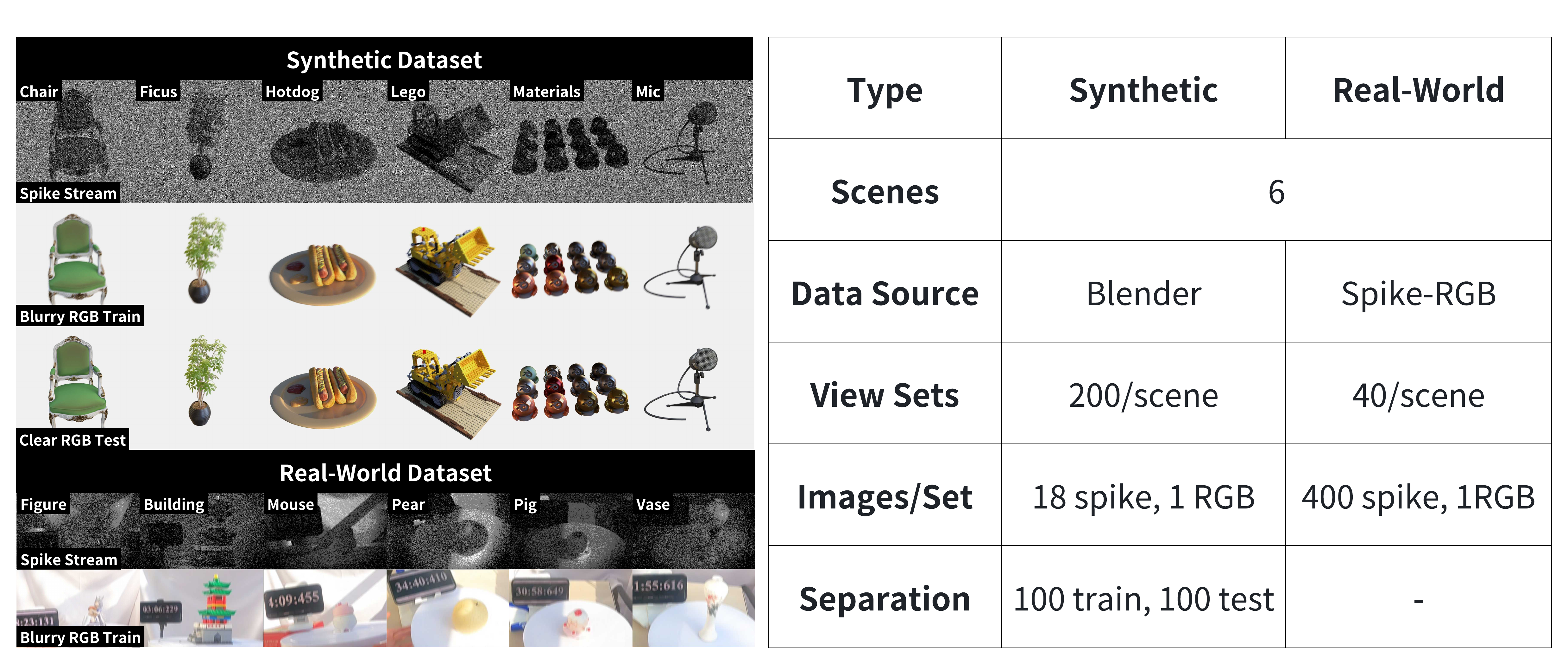}
    \caption{\textbf{Overview of the experimental setups.}  The synthetic dataset consists of spike streams, blurry RGB training data, and clear RGB test data. Relative quality metrics such as Peak Signal-to-Noise Ratio (PSNR), Structural Similarity Index Measure (SSIM), and Perceptual Similarity (LPIPS) are evaluated in the synthetic datasets. In real-world datasets where ground truth is absent, no-reference quality metrics like BRISQUE\cite{Mittal2011BlindReferencelessIS} and RankIQA\cite{Liu2017RankIQALF} are utilized. Both synthetic and real-world datasets comprise 6 unique scenes; however, the number of data varies depending on the source.}
    \label{Fig4}
\end{figure}

\paragraph{\textbf{Synthetic Dataset}}
\label{Synthetic data}
Our synthetic dataset comprises six classical scenes for NeRF reconstruction: hotdog, ficus, lego, chair, materials, and mic. We utilized Blender as our virtual environment to collect 200 sets of RGB images for each scene, all captured from different camera views. Each set consists of 18 sharp images obtained using the Camera Shakify plugin in Blender, which simulates motion blur by shaking the camera. The 200 sets of views are divided evenly into 100 training views and 100 testing views. For the training views, 18 RGB images from each view are processed through a spike-generating tool to produce corresponding binary spike streams. It is crucial to note that during the training phase, we use only a single blurry RGB image and 18 binary spike data for supervision, excluding any sharp RGB images from the process.     

\paragraph{\textbf{Real-world Dataset}}
\label{Real-world Dataset}
Based on the Spike-RGB camera system we introduced in Sec.~\ref{Spike-RGB Camera System Setup}, we constructed an RGB \& Spike 3D (RS-3D) dataset. This dataset captures six real-world scenes using the Spike-RGB camera system, with camera poses derived via COLMAP\cite{schonberger2016structure}. The creation involved capturing RGB and spike videos from 40 viewpoints, with intentional camera movement for motion blur. Temporal synchronization utilized a precise clock for accurate timestamps, aligning the spike data with RGB frames. Spatial alignment was achieved using a beam splitter and camera calibration with checkerboards, allowing for accurate cropping of RGB frames to match the spike camera's field of view. The RS-3D dataset thus includes six scenes with 40 synchronized and aligned pairs of RGB images and spike streams. Dataset details are provided in Fig.~\ref{Fig4}.

\paragraph{\textbf{Baselines}}
\label{Baseline Overview}
The aforementioned in Sec.~\ref{Introduction} highlights the existence of diverse deblur techniques, encompassing algorithm-based deblurring without reliance on hardware or sharp training data (Deblur-NeRF\cite{Ma2021DeblurNeRFNR} and MPR-NeRF\cite{Zamir2021MultiStagePI}) as well as hardware-dependent methods that employ neuromorphic cameras for capturing clear data (D2Net-NeRF\cite{Shang2021BringingEI}, EDI-NeRF\cite{Pan2018BringingAB}, and E2NeRF\cite{Qi_2023_ICCV}).
To be specific, Deblur-NeRF\cite{Ma2021DeblurNeRFNR} incorporates a deformable kernel to acquire knowledge of the blurring process, which is then passed on to NeRF MLP. MPR-NeRF\cite{Zamir2021MultiStagePI} is a single-image deblurring technique that performs image deblurring before NeRF training. D2Net-NeRF\cite{Shang2021BringingEI} and EDI-NeRF\cite{Pan2018BringingAB} employ event-based deblurring, with NeRF being trained using the resulting deblurred images. Lastly, E2NeRF\cite{Qi_2023_ICCV} integrates an event-based loss for deblurring during the NeRF training process and achieves state-of-the-art performance.

\subsection{Result Analysis}
\label{Result Analysis}

\paragraph{\textbf{Synthetic data}}
\label{Synthetic data QA}

\begin{table*}[htb]
\centering
\caption{\textbf{Quantitative analysis on blur and novel views.} The results are the averages of the six synthetic scenes. The best average score under a similar setting is marked in bold.}
\label{Tab2}
\resizebox{\textwidth}{!}{%
\begin{tabular}{|l|cc|cc|cc|}
\hline
\multirow{2}{*}{Method} & \multicolumn{2}{c|}{PSNR$\uparrow$} & \multicolumn{2}{c|}{SSIM$\uparrow$} & \multicolumn{2}{c|}{LPIPS$\downarrow$} \\ 
                        & Blur View & Novel View    & Blur View & Novel View    & Blur View & Novel View     \\ \hline
NeRF                    & 22.91     & 22.27         & .9072     & .9018         & .1441     & .1483          \\
Deblur-NeRF             & 21.71     & 19.93         & .8795     & .8584         & .2364     & .2573          \\
D2Net-NeRF              & 27.46     & 26.65         & .9450     & .9427         & .1029     & .1087          \\
EDI-NeRF                & 27.94     & 27.71         & .9497     & .9522         & .0860     & .0896          \\
MPR-NeRF                & 27.93     & 27.91         & .9525     & .9571         & .0882     & .0861          \\
E2NeRF$^{5\times}$      & 29.16     & 29.09         & .9592     & .9571         & .0828     & .0826          \\
E2NeRF$^{2\times}$      & 23.41     & 23.17         & .9048     & .9034         & .1579     & .1588          \\
SpikeNeRF$^{5\times}$   & \textbf{29.36} & \textbf{29.15 } & \textbf{.9669}  & \textbf{.9654}   & \textbf{.0625} & \textbf{.0624}  \\
SpikeNeRF               & \textbf{28.46}  & \textbf{28.27} & \textbf{.9603}  & \textbf{.9595}  & \textbf{.0787} & \textbf{.0790}  \\ \hline
\end{tabular}
} 
\end{table*}

\begin{figure*}[htb]
\centering
\includegraphics[width=\textwidth]{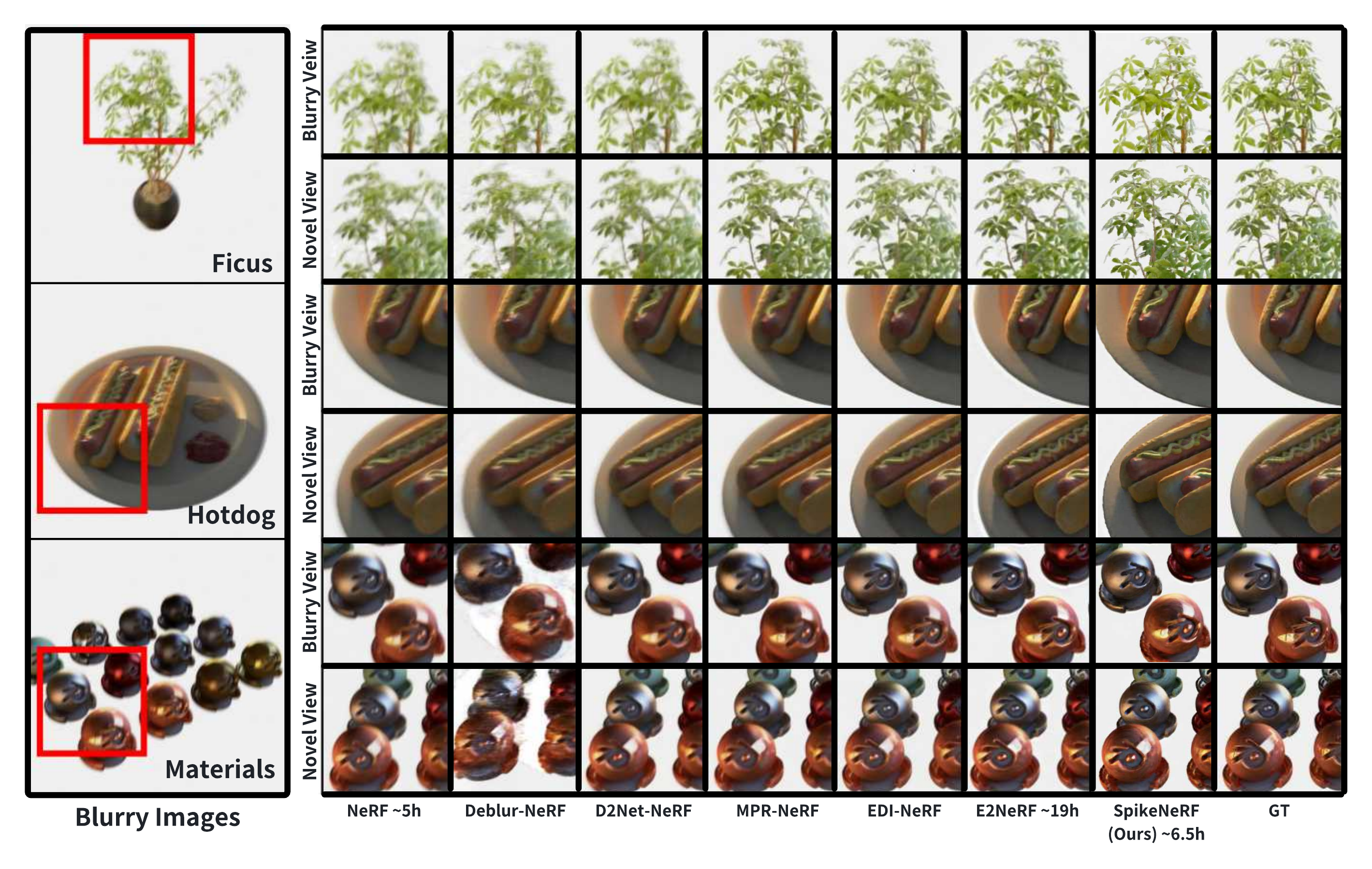}
\caption{\textbf{Visualization quality using synthetic data.} In comparison to methods like Deblur-NeRF, SpikeNeRF demonstrates superior performance due to the hardware's ability to capture sharp spike streams effectively. Moreover, SpikeNeRF exhibits efficient resource utilization by avoiding asynchronous differentiation, making it more advantageous than event-based approaches such as E2NeRF.}
\label{Fig5}
\end{figure*}

As detailed in Sec.~\ref{Baseline Overview}, we evaluated various NeRF methods using three metrics: PSNR, SSIM, and LPIPS. Our approach, integrating spike streams, outperformed baseline models across the dataset, showcasing state-of-the-art deblurring performance, and outstripping traditional NeRF deblurring techniques and E2NeRF under similar test conditions (Fig.~\ref{Fig5} and Tab.~\ref{Tab2}). 

We noted that E2NeRF necessitates independent rendering to calculate the event-based loss, incurring substantial extra computational costs. The results from the optimal settings reported in the original E2NeRF \cite{Qi_2023_ICCV}, which involved sampling 5 individual timestamps (4 bins) for each pixel (referred to as E2NeRF$^{5\times}$), were included for comparison. We also incorporated a comparable setting in SpikeNeRF$^{5\times}$. Additionally, we present results from E2NeRF's minimum cost setting (1 bin 2 individual timestamps) as E2NeRF$^{2\times}$. Lastly, the default setting of our SpikeNeRF does not entail any significant additional costs and is simply referred to as SpikeNeRF.  

In addition to quantitative analysis, we also conducted qualitative comparisons on synthetic data (Fig.~\ref{Fig5}). The experiments included the majority of other models from the quantitative analysis. From the final results, it is evident that SpikeNeRF still exhibits the best performance, reconstructing images with greater clarity and richer texture details.

\paragraph{\textbf{Real-world Data}}
\label{Real-world Dataset RA}
\begin{figure*}[htb]
\centering
\includegraphics[width=\textwidth]{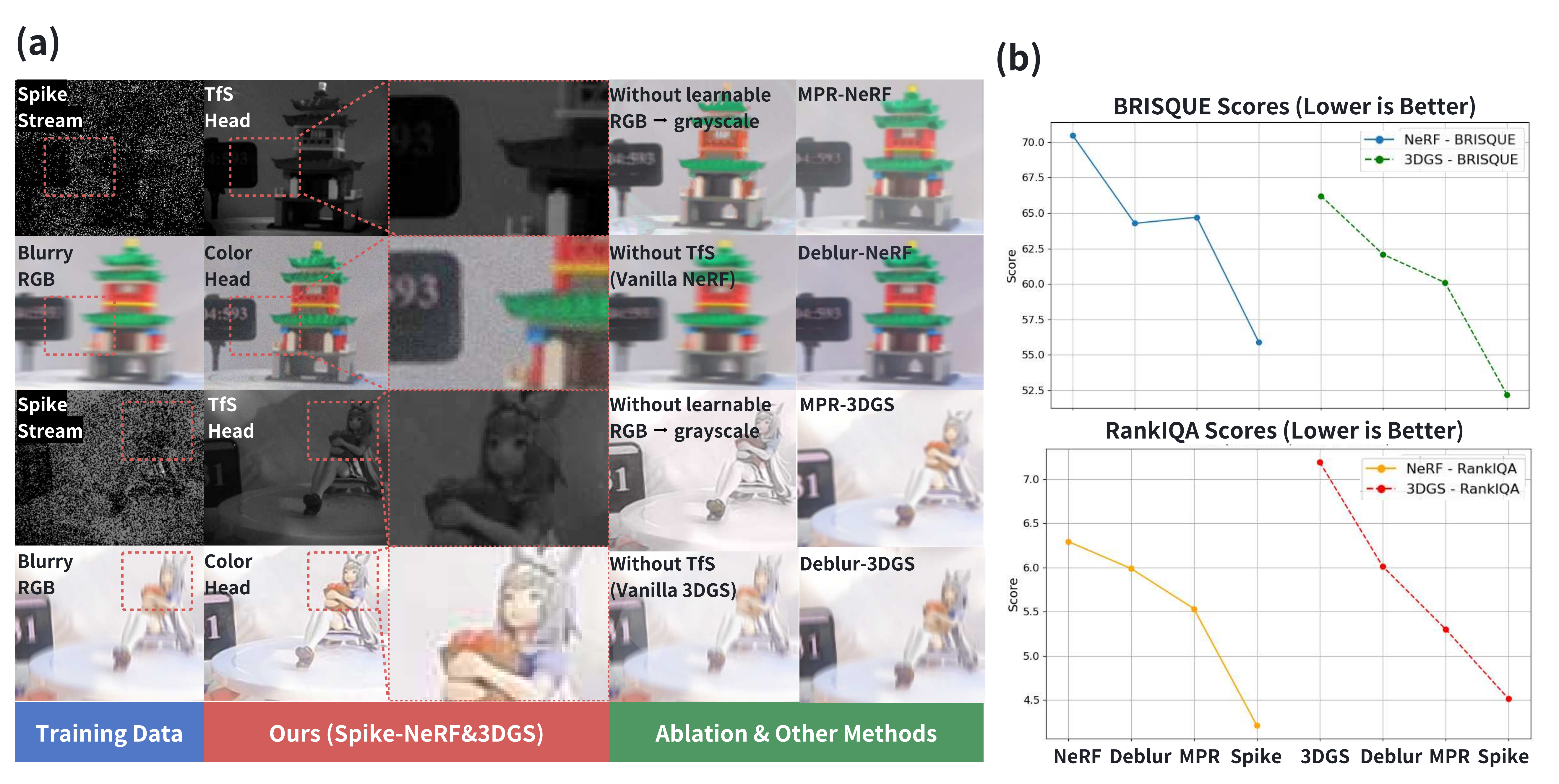}
\caption{\textbf{NeRF and 3DGS test using real-world data.} \textbf{(a)}, The visualization results demonstrate that the incorporation of our core design elements - Texture from Spike (TfS) loss and learned RGB to grayscale conversion, whether using NeRF or 3DGS, significantly enhances the final rendering outcomes. \textbf{(b)}, After considering the lack of ground truth in real data, we employed two commonly used no-reference image quality assessment metrics, namely BRISQUE\cite{Mittal2011BlindReferencelessIS} and RankIQA\cite{Liu2017RankIQALF}. Both metrics are designed with lower scores indicating better quality, and our spike-enhanced strategy yields the highest score.}
\label{Fig6}
\end{figure*}

We conducted a range of qualitative and quantitative experiments on real-world datasets, employing both NeRF and 3DGS as fundamental methodologies. Subsequently, we applied diverse deblurring techniques to process images from our RS-3D dataset, with the comparative results depicted in Fig.~\ref{Fig6}. Our approach demonstrates promising outcomes when applied to real-world data. The spike stream exhibits an exceptionally high frame rate, effectively mitigating motion blur during recorded motion processes and compensating for the loss of information inherent in the input blurry RGB frame.

\subsection{Runtime Test and Ablation Study}
\label{Ablation Study}

\begin{figure*}[hbt]
\centering
\includegraphics[width=\textwidth]{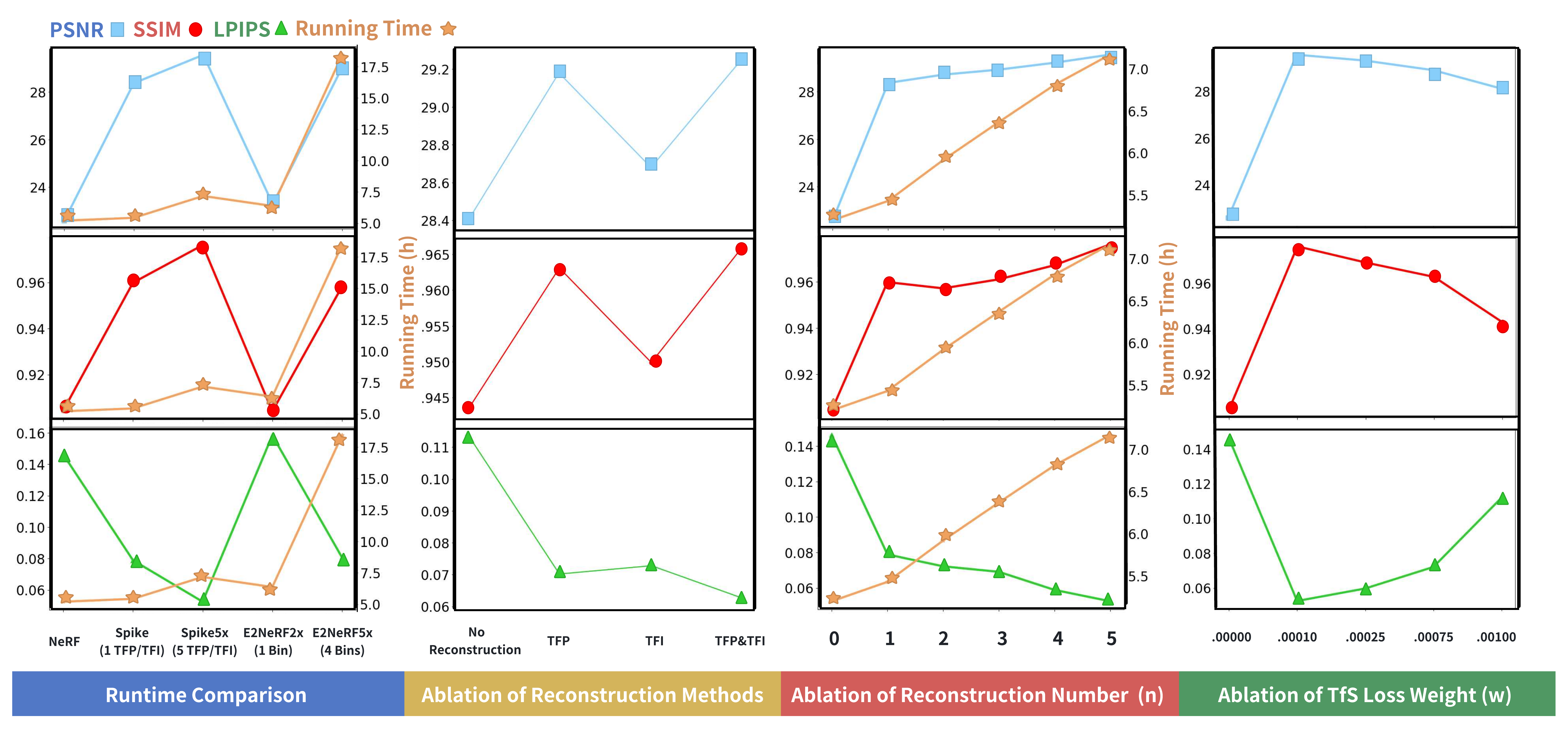}
\caption{\textbf{Further Investigation on the trade-off between cost and performance.} In the first column, we compared the performance of the spike-enhanced method (Ours) with the state-of-the-art event-enhanced method (E2NeRF\cite{Qi_2023_ICCV}). Both methods demonstrate that increasing training data and investment yields better outcomes. However, it should be noted that the minimum cost for E2NeRF is referenced as E2NeRF$^{2\times}$ and the standard cost as E2NeRF$^{5\times}$. We can achieve comparable results using only 1/4 of the resources (SpikeNeRF vs E2NeRF$^{5\times}$) and significantly outperform when matching the number of training data samples (SpikeNeRF$^{5\times}$ vs E2NeRF$^{5\times}$). To further explore a better balance between cost and performance, we conducted ablation experiments on different spike reconstruction techniques, varying the number of reconstructions per camera view, and adjusting the weighting allocated to TfS loss (columns 2-4).
The results indicate that even a single spike reconstruction from both TFP and TFI can yield excellent outcomes under relatively small weightings (0.0001).
}
\label{Fig7}
\end{figure*}
The rationale behind the superior efficiency of spike streams over event streams in learnable 3D reconstruction tasks lies primarily in their ability to avoid independent rendering along the time axis. Experimental results from Fig.~\ref{Fig7} corroborate our hypothesis. 
In addition to conducting runtime experiments, we also investigated the impact of different spike texture reconstruction methods. Specifically, we compared experiments involving direct utilization of raw spike streams, TFP texture, TFI texture, and a combination of both. From the results obtained, it is evident that employing both TFP and TFI yields benefits across all metrics, particularly perceptual loss (LPIPS). Subsequently, we conducted experiments with varying scales for the number of spike texture reconstructions used in training each view ($n$) and weight ($w$) of the TfS loss. Upon reaching a training dataset size of 1 for spike data, there was a noticeable plateau in metric growth indicating that spike-enhanced deblur exhibits higher data efficiency compared to the event method. The choice of TfS loss weight ($w$) had a subtle impact on experimental outcomes; within our settings, its effect peaked at an optimal value (0.0001).

\section{Potential Advantages}
\label{Conclusion and Future Works}

\begin{figure*}[htb]
\centering
\includegraphics[width=\textwidth]{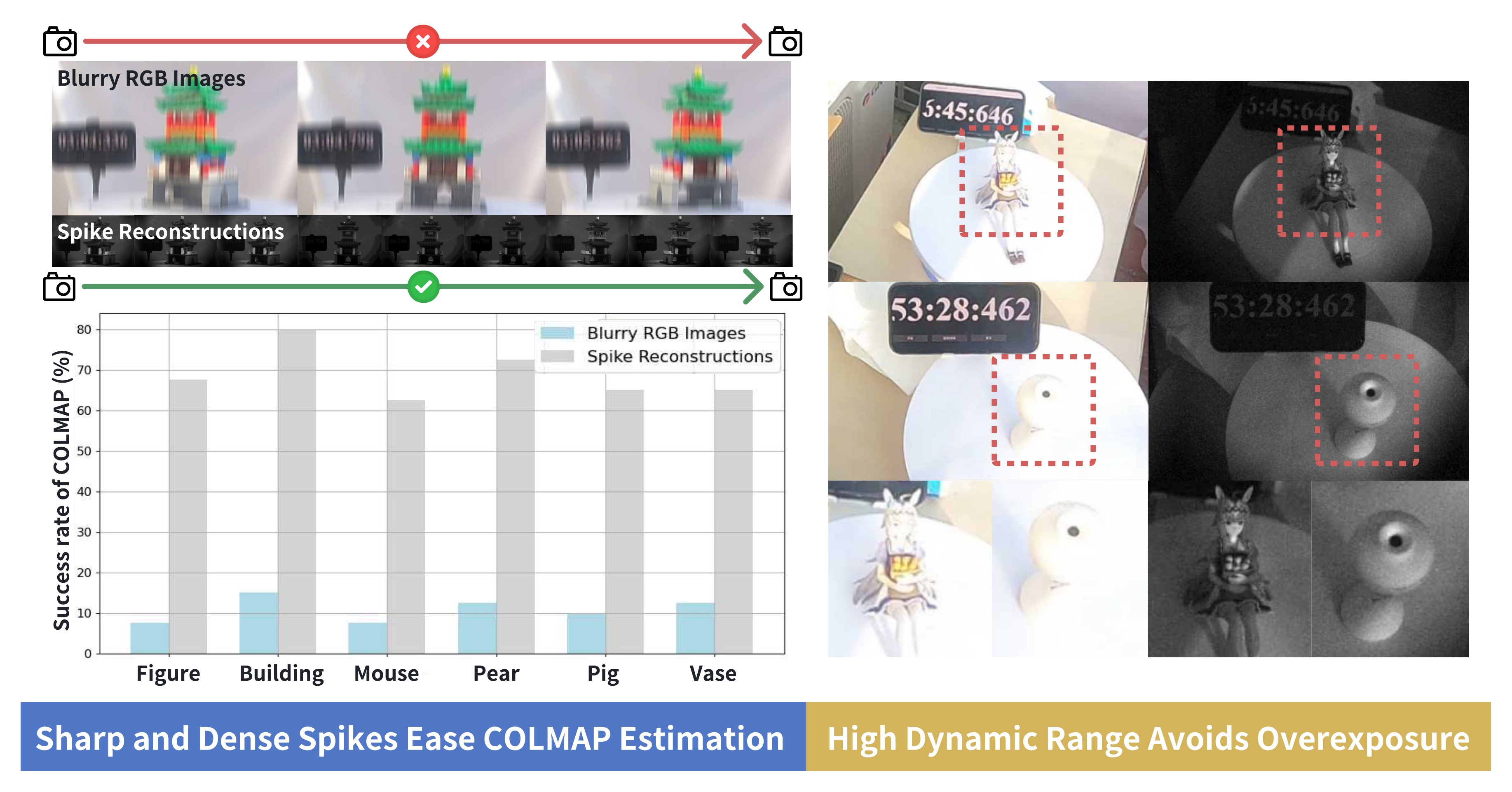}
\caption{\textbf{Delve deeper into the potential advantages of the spike-enhanced method.} The left column reveals that in addition to our primary focus on the efficacy of spike streams in enhancing learning-based 3D reconstruction deblurring, they also provide a more precise camera pose estimation by offering richer temporal reference. In our real-world dataset, spike streams elevate the success rate of COLMAP by fourfold. Furthermore, we observe that spike cameras exhibit a higher dynamic range, thereby mitigating overexposure (right column).
}
\label{Fig8}
\end{figure*}

In Fig.~\ref{Fig8} we demonstrate additional advantages that spike streams bring to our test, including facilitating camera pose estimation and mitigating overexposure. We utilize COLMAP \cite{schoenberger2016sfm, schoenberger2016mvs} for camera pose estimation. Direct estimation of camera poses from blurry RGB inputs is notoriously difficult. However, our experiments reveal that spike reconstruction significantly enhances COLMAP's ability to accurately estimate camera poses. The robustness of COLMAP's model estimation algorithms, such as RANSAC and bundle adjustment, largely depends on the availability of high-quality and consistent feature points across images. Spike reconstruction contributes to this process by providing denser and clearer feature points, which are essential for precise feature matching in COLMAP.

\section{Conclusion and Future Works}
\label{Conclusion and Future Works}
In summary, the findings of our study showcase the superior performance of spike streams compared to event streams in the context of 3D reconstruction with blurry images task from various perspectives.  Our novel 'Texture from Spike' (TfS) loss function, coupled with an efficient end-to-end training process, demonstrates state-of-the-art performance without incurring substantial additional computational costs. Besides, the creation of the synthetic and real-world RS-NeRF datasets marks a pioneering step towards furthering research in the mutual enhancement of spike data and Implicit Neural Representation (INR) networks. Our approach opens new avenues for leveraging the high temporal resolution of spike cameras, surpassing traditional RGB cameras in capturing fast-moving scenes without motion blur and in challenging lighting conditions.

Future work could explore the extension of the SpikeNeRF framework to other domains where high-speed capture is crucial. Despite the limited availability of spike hardware in the market, we anticipate that future research will unveil further applications for spikes in diverse visual tasks, similar to event cameras but with their distinctive attributes.

\clearpage  

\clearpage
\setcounter{page}{1}
\setcounter{section}{0}

\title{Appendix - SpikeNeRF: Enhancing Learning-based 3D Reconstruction from Blurry Images using Spike Camera}
\maketitle

\section{Overview}
\label{sec: Overview}

This appendix of SpikeNeRF includes:

\begin{itemize}
    \item \textbf{Experimental Setups:} The model aligns with standard NeRF configurations in terms of depth and width parameters, trained on a single NVIDIA A100 GPU. Baseline description and detailed hyperparameters are provided, ensuring a balance between computational efficiency and performance.
    \item \textbf{Computational Complexity Analysis:} A detailed comparison between the computational costs of E2NeRF and SpikeNeRF is provided, highlighting the efficiency of SpikeNeRF in resource management.
    \item \textbf{Comparison between Cameras and Future works} The cost of the equipment poses a practical concern. We conducted a comparison of the cost and configuration between spike, event, and high-speed RGB cameras. Additionally, we explored the potential applications of next-generation spike cameras that incorporate an RGB channel.
\end{itemize}

\section{Experimental Setups}
\label{sec: Experimental Setups}

Our study undertakes a detailed comparison with the E2NeRF model, building upon its framework and incorporating elements from the standard NeRF architecture as described in \cite{Qi_2023_ICCV}. The training was conducted on a single NVIDIA A100 GPU, mirroring the typical resource allocation in standard NeRF implementations. Notably, our approach aligns with common NeRF configurations, specifically in terms of the `depth` and `width` parameters of the model. These parameters are crucial, as the `depth` corresponds to the number of `Dense` layers, and the `width` to the number of units in each layer, as is standard in NeRF models. We also includes the baseline description in Tab.~\ref{Tab4}
\begin{table}[ht]
\centering
\caption{Detailed Hyperparameters}
\begin{tabular}{lc}
\hline
\textbf{Parameter} & \textbf{Value} \\ \hline
Weight ($w$)         & 0.0001                  \\
Threshold ($\Omega$) & 2                      \\     
Batch Size           & 1024                   \\ 
Coordinate Normalization Range & [-1, 1]      \\ 
TFP Window Size      & 6                      \\
Iteration Count      & 200,000               \\ \hline
\label{Tab3}
\end{tabular}
\end{table}

\begin{table}[htb]
\centering
\caption{
The deblur-NeRF\cite{Ma2021DeblurNeRFNR} incorporates a deformable kernel to acquire knowledge of the blurring process, which is then passed on to NeRF MLP. MPR-NeRF\cite{Zamir2021MultiStagePI} is a single-image deblurring technique that performs image deblurring before NeRF training. D2Net-NeRF\cite{Shang2021BringingEI} and EDI-NeRF\cite{Pan2018BringingAB} employ event-based deblurring, with NeRF being trained using the resulting deblurred images. Lastly, E2NeRF\cite{Qi_2023_ICCV} integrates an event-based loss for deblurring during the NeRF training process and achieves state-of-the-art performance.
}
\begin{tabular}{|l|p{0.7\columnwidth}|}
\hline
\textbf{Method}        & \textbf{Description}                          \\ \hline
Deblur-NeRF\cite{Ma2021DeblurNeRFNR}            & Deblur with extra network, NeRF train on blurry images   \\ \hline
MPR\cite{Zamir2021MultiStagePI}-NeRF               & Single-image deblurring, NeRF train on deblurred images \\ \hline
D2Net\cite{Shang2021BringingEI}-NeRF                 & Event-based deblurring, NeRF train on deblurred images \\ \hline
EDI\cite{Pan2018BringingAB}-NeRF                & Event-based deblurring, NeRF train on deblurred images \\ \hline
E2NeRF\cite{Qi_2023_ICCV}                 & Event-based loss deblur during training NeRF      \\ \hline
\end{tabular}
\label{Tab4}
\end{table}



\section{Analysis of Complexity: Computational Cost Comparison Between E2NeRF and SpikeNeRF}
\label{sec: Complexity Analyze:}
As we discussed in Sec. 1 of our main text, there exists a fundamental trade-off between event loss accuracy and computational efficiency in event-based neural rendering frameworks. To illustrate, consider two sampled timestamps $t_{0}$ and $t_{1}$, which exhibit similar brightness values. These values do not trigger an event as their difference falls within the non-activation range ($-\Theta < b_{t_{1}} - b_{t_{0}} < \Theta$), where $\Theta$ denotes the event threshold. However, introducing an intermediary timestamp $t_{\frac{1}{2}}$ reveals that an event does occur within the interval $t_{0}$ to $t_{1}$, particularly if $b_{t_{\frac{1}{2}}} - b_{t_{0}} > \Theta$. 

A straightforward solution to this detection dilemma is to decrease the time sampling interval, thereby capturing more granular changes in brightness. Yet, this approach introduces a significant computational overhead. In E2NeRF \cite{Qi_2023_ICCV}, for instance, each additional sampled timestamp proportionally increases the computational load for event-based loss computation, as illustrated in Fig. 1 in our main text.

\begin{figure*}[htb]
\centering
\includegraphics[width=\textwidth]{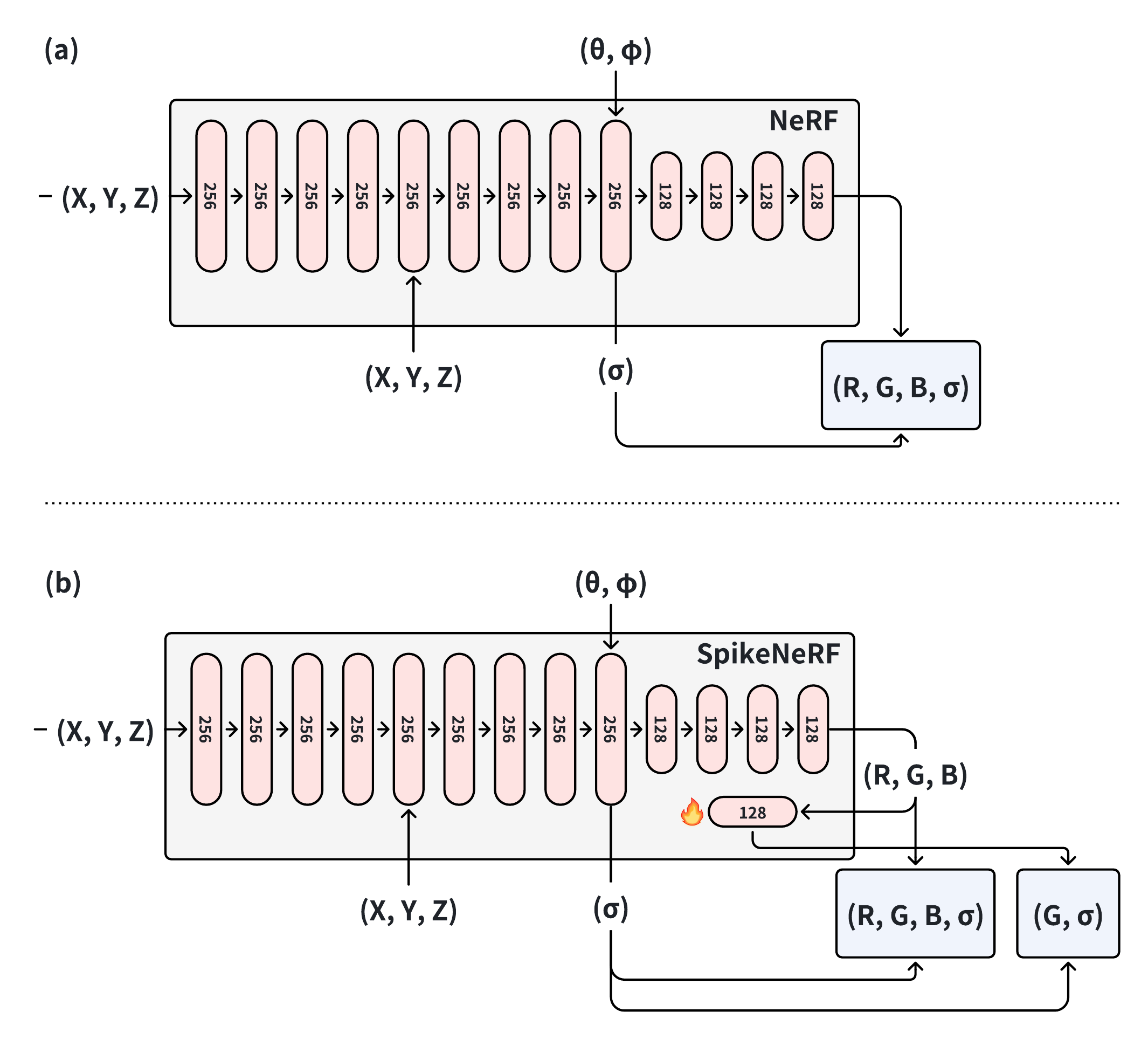}
\caption{Comparative Architecture of NeRF and SpikeNeRF.}
\label{fig: structure}
\end{figure*}

To quantify this complexity, we analyze the inference operations in the original NeRF network architecture, as shown in Fig.~\ref{fig: structure} (a). The computational load for each fully connected (dense) layer is determined by the number of neurons in both the current and preceding layers. Consequently, the total inference complexity, denoted as $C$, is calculated by summing the products of neuron counts across successive layers:
\begin{equation}
C = n_{0} \cdot n_{1} + n_{1} \cdot n_{2} + \cdots + n_{L-2} \cdot n_{L-1} + n_{L-1} \cdot n_{L}
\label{Eq.11}
\end{equation}
In the case of E2NeRF, which employs an event-based loss, each rendered event requires independent calculations, resulting in a computational cost of at least $N \times C$ if we have $N$ sampling timestamps. Conversely, SpikeNeRF primarily incurs additional computational expenses in the RGB converter layer, represented by $O(n_{L-1} \cdot n_{L})$. This distinction underscores the efficiency of SpikeNeRF in managing computational resources.

\section{Comparison between Cameras and Future works}
\begin{figure*}[htb]
\centering
\includegraphics[width=\textwidth]{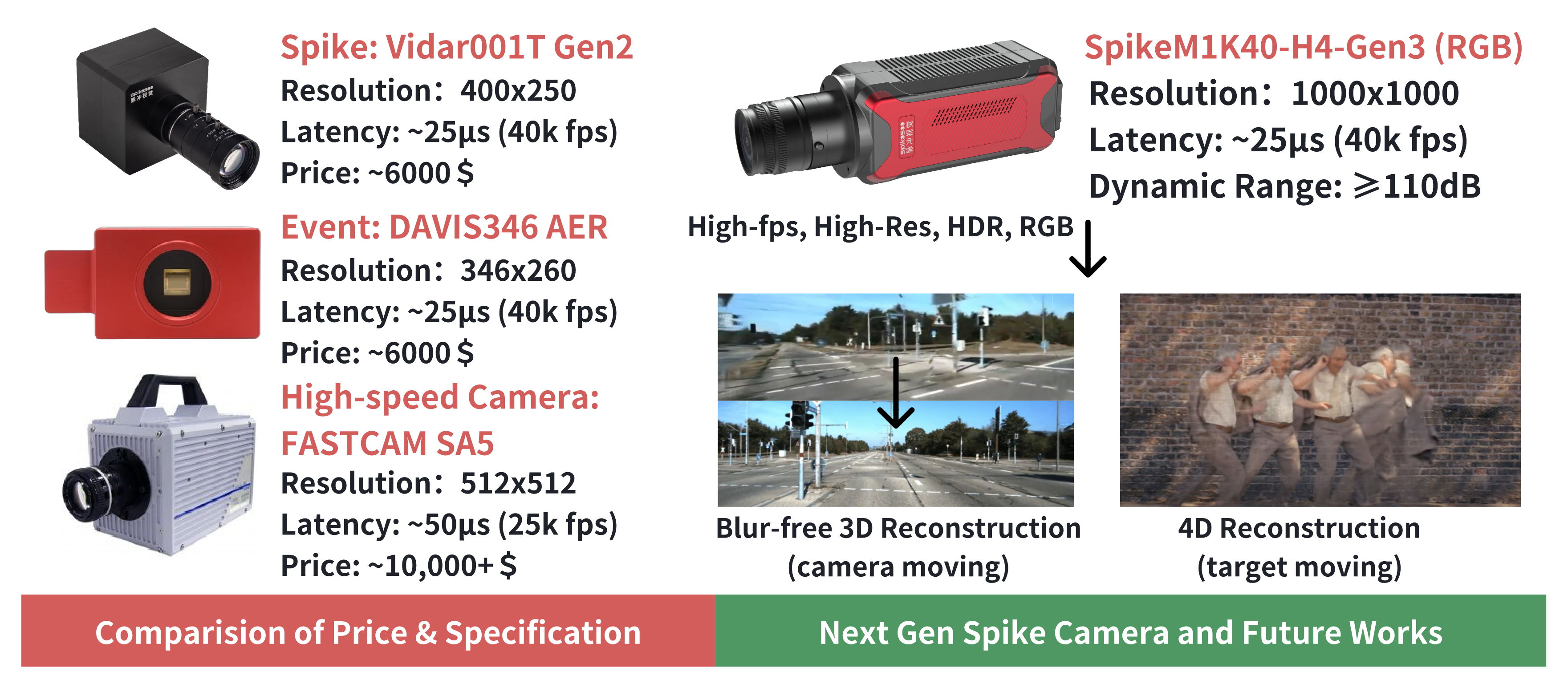}
\caption{The right column presents a comparative overview of price and configuration between spike, event, and regular high-speed cameras. The left column showcases the potential applications of the next generation spike camera.}
\label{fig:cost}
\end{figure*}

 In Fig.~\ref{fig:cost} Column 3 presents a comparison between spike cameras, event cameras, and regular high-speed cameras in terms of specifications and prices. Spike cameras prove to be more cost-effective than high-speed cameras; with similar hardware costs, they yield superior reconstruction results while requiring significantly less training expenditure compared to event cameras. In column 4, next-generation spike cameras offer higher resolution and RGB content capabilities, enabling direct utilization for NeRF or 3DGS training purposes—a potential breakthrough when considering scenes involving relative movement between the camera and targets.

\bibliographystyle{splncs04}
\bibliography{main}

\end{document}